\documentclass{article} % For LaTeX2e
\usepackage{iclr2019_conference,times}

\usepackage{amsmath,amsfonts,bm}

\def\eqref#1{equation~\ref{#1}}
\def\1{\bm{1}}

\DeclareMathAlphabet{\mathsfit}{\encodingdefault}{\sfdefault}{m}{sl}
\SetMathAlphabet{\mathsfit}{bold}{\encodingdefault}{\sfdefault}{bx}{n}

\usepackage{booktabs}
\usepackage{url}
\usepackage{graphicx}
\usepackage{amssymb}

\usepackage[T1]{fontenc}
\usepackage[utf8]{inputenc}

\usepackage{wrapfig}

\usepackage{enumitem}

\newcommand*\samethanks[1][\value{footnote}]{\footnotemark[#1]}

\iclrfinalcopy

\title{Simulating Execution Time of Tensor Programs using Graph Neural Networks}

\author{Jakub M. Tomczak\thanks{All authors contributed equally.}\ \ , Romain Lepert\samethanks \ \ \& Auke Wiggers\samethanks \ \ \\
Qualcomm AI Research\thanks{Qualcomm AI Research is an initiative of Qualcomm Technologies, Inc.}\\
Qualcomm Technologies Netherlands B.V. \\
\texttt{\{jtomczak, romain, auke\}@qti.qualcomm.com} \\
}

\newcommand{\eg}{\textit{e.g.}}
\newcommand{\ie}{\textit{i.e.}}

\begin{document}

\maketitle

\begin{abstract}
    Optimizing the execution time of tensor program, \eg, a convolution, involves finding its optimal configuration.
	Searching the configuration space exhaustively is typically infeasible in practice.
	In line with recent research using TVM, we propose to learn a surrogate model to overcome this issue. 
	The model is trained on an acyclic graph called an abstract syntax tree, and utilizes a graph convolutional network to exploit structure in the graph.
	We claim that a learnable graph-based data processing is a strong competitor to  heuristic-based feature extraction.
 	We present a new dataset of graphs corresponding to configurations and their execution time for various tensor programs.
	We provide baselines for a runtime prediction task.
\end{abstract}

%--SECTION--
\section{Introduction}
Current deep learning frameworks, such as TensorFlow, PyTorch, allow to optimize a computational graph representation using, \eg, auto differentiation and memory management \citep{abadi2016tensorflow, paszke2017automatic}. 
However, they do not tackle optimization of hardware-specific operator-level transformations, but rely on manually tuned and vendor-specific operator libraries. 
Thus, there is room to further improve a computational graph by optimizing transformations for specific hardware.

Recently, this gap has been filled by TVM, a compiler framework that allows both graph- and operator-level optimization in an end-to-end manner \citep{chen2018tvm}. 
TVM specifies a \textit{configuration} for an operator, \eg, a specific way of performing a convolution, and compiles the resulting \textit{tensor program} to a target hardware.
%The main challenge in runtime optimization is the dependency of the compiler on the operation and the target hardware. 
As a consequence, for each new workload/operator, optimization over a new configuration space must be carried out.
This results in a hard optimization problem, \eg, for Nvidia GPU the search space of a single operator consists of more than $10^6$ configurations.

Recent efforts overcome this issue by learning how to optimize tensor programs from data \citep{chen2018learning}. 
Instead of running an exhaustive search over an impractically large search space, a \textit{surrogate model} is trained to predict runtime for a given configuration.
This model is in turn used to select the configuration that minimizes the runtime.
\citep{chen2018learning} utilizes XGBoost \citep{chen2016xgboost} and TreeGRU \citep{kai2015improved} as surrogate models.

%\todo{AW:Learning to optimize tensor programs does all this and more, we should differentiate from them here.. we train on small data, test found representations instead of optimizing tensor programs}
%In this paper we follow this line of thinking and propose a surrogate model architecture based on Graph Neural Networks (GraphNNs).
%\todo{AW: all of these advantages also work for chen2018, should we perhaps t-sne our found representations}
%This serves the following purposes. 
%First, training a simulator allows to quickly predict next promising configurations for a hardware. 
%Querying the surrogate model takes miliseconds for a single configuration, which is two or more orders of magnitudes faster than compiling and measuring the configuration on the target hardware.
%Second, learning a model allows to transfer it across operators/workloads \citep{chen2018learning}. 
%Third, we can also use the simulator for verifying optimization techniques and fast prototyping.

%In particular, we investigate the usefulness of GraphNNs on a small dataset.

\paragraph{Contribution}
Similar to \citep{chen2018learning}, we represent a configuration of a tensor operator as an abstract syntax tree (AST) \citep{allamanis2017learning}, and extract node features using TVM. 
We then train a Graph Neural Network (GraphNN) on the resulting graph as the surrogate model. 
%\ie, one or more graph-convolutional layers followed by dense layers, 
We claim that GraphNNs are a good fit, as, crucially, they preserve the graph structure of the AST and allow propagating information among nodes.
The contribution of the paper is threefold:
\begin{itemize}[leftmargin=*]
\item We present a new problem for GraphNNs: predicting the execution time of tensor programs from their corresponding AST. 
For this purpose, we gathered a \textbf{new dataset} and we propose to use it as a new application in the GraphNN community.\footnote{The dataset can be downloaded at: \url{https://developer.qualcomm.com/project/qast}} This is the main contribution of the paper.
\item We propose to use a graph neural network as a surrogate model of a compiler. We claim that it is important to use a learnable graph data transformation rather than a fixed feature extractor, \eg, context relation features \citep{chen2018learning}.
\item We perform experiments on the newly proposed dataset and provide baseline results for a cross-workload prediction task.
\end{itemize}

\paragraph{Related work} 
GraphNNs have been proven to be powerful in many applications ranging from chemistry and life sciences \citep{de2018molgan, duvenaud2015convolutional, gonczarek2018interaction, zitnik2018modeling} to social networks \cite{davidson2018hyperspherical, kipf2016semi, hamilton2017representation, velivckovic2017graph}, where graph inputs represent chemical compounds or social interactions among users. 
GraphsNNs find applications in other domains as well, such as geometric modeling \citep{bronstein2017geometric}, recommendation systems \citep{berg2017graph}, relational data and knowledge graphs \citep{nickel2016holographic, schlichtkrull2018modeling}, and regression problems like chemical properties prediction \citep{duvenaud2015convolutional, li2018adaptive} and traffic prediction \citep{yu2017spatio}.
An interesting on-going research is on using these networks in generative settings \citep{de2018molgan, jin2018junction, simonovsky2018graphvae}.

\citep{chen2018learning} introduced the idea of learning a surrogate model using the TVM framework.
We instead use GraphNNs in this context, and do not focus on runtime minimization yet.
%\todo{AW: highlight that we're focusing on it in future work or something?}

%--SECTION--
%\section{Dataset}
%\input{dataset.tex}

%--SECTION--
%\section{Methodology}
\begin{figure}[!tp]
\centering
\includegraphics[width=0.9\textwidth]{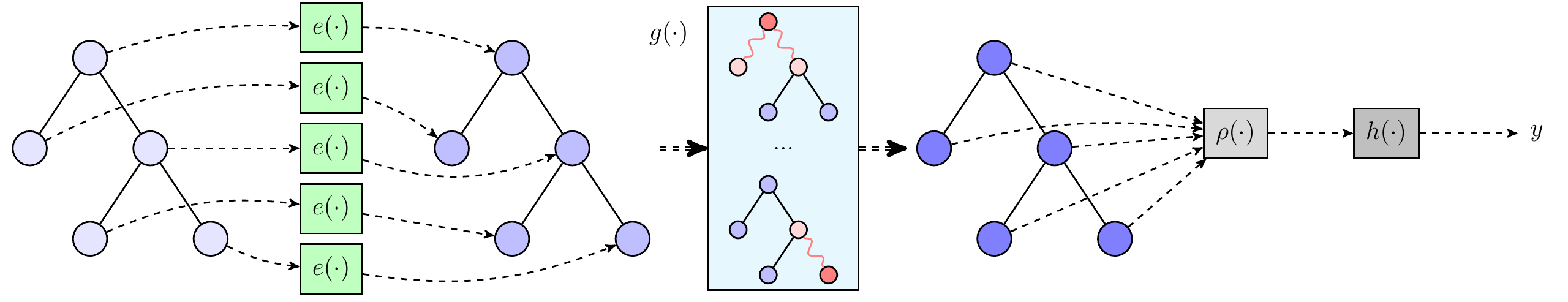}
\vskip -2  mm
		\caption{A schematic presentation of our approach: First, each node is transformed by a shared encoder (green rectangle), then a graph convolutional network is used to propagate information between nodes (cyan rectangle). Finally, all nodes are aggregated and a prediction $y$ is made.}
\label{fig:approach}
\end{figure}

%---SECTION---
\section{Problem formulation} 

We consider the problem of learning an \textit{execution time simulator}.
That is, we aim to predict runtime of a configuration $x \in \mathcal{X}$ on the target hardware, where $\mathcal{X}$ is a discrete configuration space. Importantly, we represent $x$ as a graph corresponding to the abstract syntax tree (AST) of the configuration.
Let us define the runtime as a function $y = f(x)$, $y \in \mathbb{R}_{+}$. 
The function $f$ can be queried, however, its analytical form is unknown. Our goal is to learn a surrogate model of the function $f$, denoted by $\hat{f}$, such that we minimize a loss function $\ell(y,\hat{f}(x))$ (e.g., $\ell_{1}$-norm or, equivalently, the logarithm of the Laplacian distribution). 
Assuming a collection of measurements $\mathcal{D}=\{(x_{n}, y_{n})\}_{n=1}^{N}$, a straightforward way to learn $\hat{f}$ is to minimize the objective $\mathcal{L}(\hat{f};\mathcal{D}) = \frac{1}{N} \sum_{n=1}^{N} \ell\big{(} y_n, \hat{f}(x_{n}) \big{)}$.

%\todo{AW: this is overly complicated IMO, no need to be this formal, also we should perhaps introduce terminology? e.g. Workload? }
%Let us consider a tensor operator using index expressions where the space of index expressions is denoted by $\mathcal{E}$. 
%For given expression $e \in \mathcal{E}$, we define the space of possible transformations (schedules) for $e$ by $\mathcal{S}_{e}$. 
%Further, we denote a compiler framework as a function $g: \mathcal{S}_{e} \rightarrow \mathcal{X}$, $x = g(s;e)$, where $x$ is a low-level abstract syntax tree (AST). 

%---SECTION---
\section{Dataset}
\begin{wrapfigure}{r}{7cm}
		\vspace{-0.6cm}	
		\centering
		\includegraphics[width=0.35\textwidth]{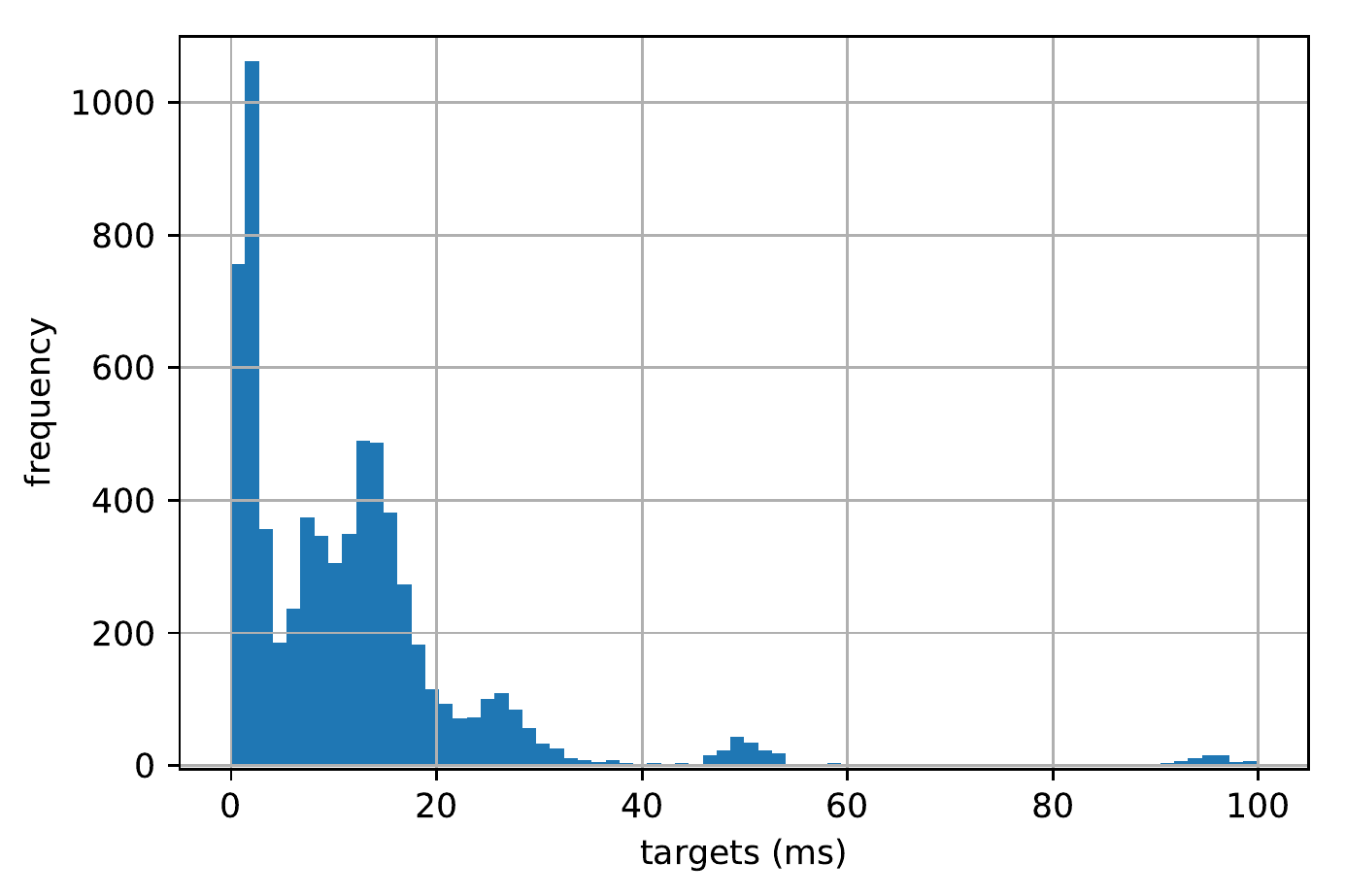}
		\vspace{-0.5cm}
		\caption{The distribution of targets.}
		\label{fig:targetdistribution}
	\end{wrapfigure}
	
\paragraph{General information}
We collect data for the operators defined in a ResNet18 \citep{he2016deepresidual}.
The target hardware is an Intel Xeon CPU E5-1620 v4 processor @ 3.50GHz, with x86-64 instruction set.
The ResNet18 architecture defines twelve unique convolution workloads, \ie, there are twelve parameterizations for the convolution operators in the network.
We show these in Table \ref{tab:resnetworkload}.
For example, for workload 'C1', the convolution has 3 input feature maps of size $224 \times 224$, and has 64 output channels.
We also show the number of configurations (\# configs) for each workload for the x86 target hardware. We performed all measurements using TVM v0.5 \citep{chen2018tvm}.

%\vspace{-0.5cm}
\begin{table}[!htp]
	\caption{Parameters for each unique Conv2D workload in a ResNet18 for x86 CPU target.}
	\label{tab:resnetworkload}
	\vspace{-0.1cm}
	\begin{center}
	\resizebox{0.75\textwidth}{!}{
	\begin{tabular}{c | r r r r r r r r | c}
		Workload & H   & W   & C$_{in}$ & C$_{out}$ & kernel & stride & padding & dilation & \# configs \\ 
\hline
    	C1    & 224 & 224 & 3        & 64        & (7, 7) & (2, 2) & (3, 3)  & (1, 1)   & 252 \\
    	C2    &  56 &  56 & 64       & 64        & (3, 3) & (1, 1) & (1, 1)  & (1, 1)   & 784 \\
    	C3    &  56 &  56 & 64       & 64        & (1, 1) & (1, 1) & (1, 1)  & (1, 1)   & 784 \\
    	C4    &  56 &  56 & 64       & 128       & (3, 3) & (2, 2) & (1, 1)  & (1, 1)   & 672 \\
    	C5    &  56 &  56 & 64       & 128       & (1, 1) & (2, 2) & (1, 1)  & (1, 1)   & 672 \\
    	C6    &  28 &  28 & 128      & 128       & (3, 3) & (1, 1) & (1, 1)  & (1, 1)   & 768 \\
    	C7    &  28 &  28 & 128      & 256       & (3, 3) & (2, 2) & (1, 1)  & (1, 1)   & 576 \\
    	C8    &  28 &  28 & 128      & 256       & (1, 1) & (2, 2) & (1, 1)  & (1, 1)   & 576 \\
    	C9    &  14 &  14 & 256      & 256       & (3, 3) & (1, 1) & (1, 1)  & (1, 1)   & 648 \\
    	C10   &  14 &  14 & 256      & 512       & (3, 3) & (2, 2) & (1, 1)  & (1, 1)   & 360 \\
    	C11   &  14 &  14 & 256      & 512       & (1, 1) & (2, 2) & (1, 1)  & (1, 1)   & 360 \\
    	C12   &   7 &   7 & 512      & 512       & (3, 3) & (1, 1) & (1, 1)  & (1, 1)   & 400 \\
	\end{tabular}
	}
	\end{center}
\end{table}
%\vspace{-0.6cm}

\paragraph{Feature representation}
The dataset contains 6,852 configurations. 
For each configuration, we extract the corresponding AST that is represented by: an adjacency matrix $A$, a feature matrix $F$, the node types (\eg, \texttt{for} statement, hardware instructions) $G$ for every node in the graph.
We save these matrices and the corresponding measured execution time $y$ as tuples $(A, F, G, y)$. The feature extraction procedure follows the one for loop context features presented in \citep{chen2018learning}.

\paragraph{Distribution of target runtimes}
The distribution of execution times do not match a normal distribution, but instead resembles a mixture of Gaussians (see Figure \ref{fig:targetdistribution}).
This poses two important challenges: (i) the simulator needs to learn a representation for all modes, (ii) generalizing to components of low probability is troublesome if these configurations correspond to unseen workload.

%	\begin{figure}[!tp]
%		\caption{The distribution of runtimes in the dataset. PLACEHOLDER}
%		\label{fig:targetdistribution}
%		\includegraphics[width=0.4\textwidth]{dataset_targets.png}
%	\end{figure}

%---SECTION---
\section{Graph neural networks as a simulator}

%We treat the input to the simulator as the AST, \ie, a graph input. 
%We determine features in nodes of the AST using a feature extraction proposed in \cite{chen2018learning}. 
%Then, our main goal is to deal with a graph input. 
In general, we can build a simulator $\hat{f}$ in a similar manner as it is presented in \citep{zaheer2017deep}: $\hat{f}(x) = h\Big{(} \rho \big{[} g\big{(} e(x_{1}), e(x_{2}), \ldots, e(x_{2}) \big{)} \big{]} \Big{)}$.
The surrogate model $\hat{f}$ consists of four components:
\begin{itemize}[leftmargin=*]
	\item A function $e(\cdot)$ that \textit{encodes} each node represented by raw features to a vector of a fixed size. Weights of the encoder are shared across nodes.
	\item A \textit{feature propagation function} $g(\cdot)$ that ensures feature information is propagated across nodes. 
	\item An \textit{aggregation function} $\rho[\cdot]$ that combines information from all nodes into a fixed-size vector. 
		Typically, it is implemented using \textit{sum} or \textit{mean}.
	\item A \textit{prediction function} $h(\cdot)$ that takes as input the fixed-size aggregated feature vector and predicts a scalar runtime $y$.
\end{itemize}

See Figure \ref{fig:approach} for a visual representation of the surrogate model (the runtime simulator).
Note that while we implement the functions above as neural networks in order to ensure that the surrogate model can be trained end-to-end using backpropagation, the four-stage framework allows us to plug in other functions as well, as long as in- and output-constraints are met. 
Importantly, the propagation function is implemented using graph convolutions \citep{duvenaud2015convolutional, kipf2016semi}.

%--SECTION--
\section{Experiments}
\subsection{Setup} 

We investigate two classes of network architectures on the extracted dataset.
The first class consists of fully-connected layers and does not propagate information among nodes before aggregation. We refer to this approach as MLP.
The second class adds the propagation of the information using a graph convolutional network (GCN).
In both classes, the \text{encoding function} $e(\cdot)$ and the prediction function $h(\cdot)$ are multi-layer perceptrons with ReLU activation.
In all models we choose the aggregation operation $\rho[\cdot]$ to be the average.
Additionally, both network types encode the \textit{node type} using a learned embedding of 32-dimensional vectors, and concatenate the resulting embedding vector to the node features.
The prediction function is composed of two fully-connected layers and the final layer has no activation function.
Finally, the prediction function is composed of two fully-connected layers and the final layer has no activation function.
For comparison, we also run a fixed feature extractor, namely, \textit{context relation features} \citep{chen2018learning}, with a learnable predictor on top. We refer to this approach as ''Curve''.
For more details, please see Table \ref{tab:architectures} in the supplementary material.
During training we use the Huber loss that is a composition of the $\ell_1$ and $\ell_2$ losses, and $\ell_1$ for final evaluation.
We use DGL package \citep{dgl} for implementing graph convolutions.

We aim at evaluating the \textbf{cross-workload generalization} capability of the surrogate model.
Such a scenario corresponds to a real-life situation where we do not have access to all workloads.
In the experiment, workloads \{C1, C2, C4, C8, C9, C12\} are taken as given data and the remaining workloads are used as test data.
The rationale is to pick as varied workloads as possible.
We split given configurations into $80\%$ of training data and $20\%$ of validation data.
Further, we consider a couple of cases where we have a limited access to training data, namely, we learn models on $5\%$, $10\%$, $15\%$, $20\%$, $25\%$, $50\%$, $75\%$ and, for completeness, $100\%$ of the training set.
This limitation imitates the real-life situation where we do not have measurements for all configurations.

\begin{figure}[!tbp]
\includegraphics[width=0.45\textwidth]{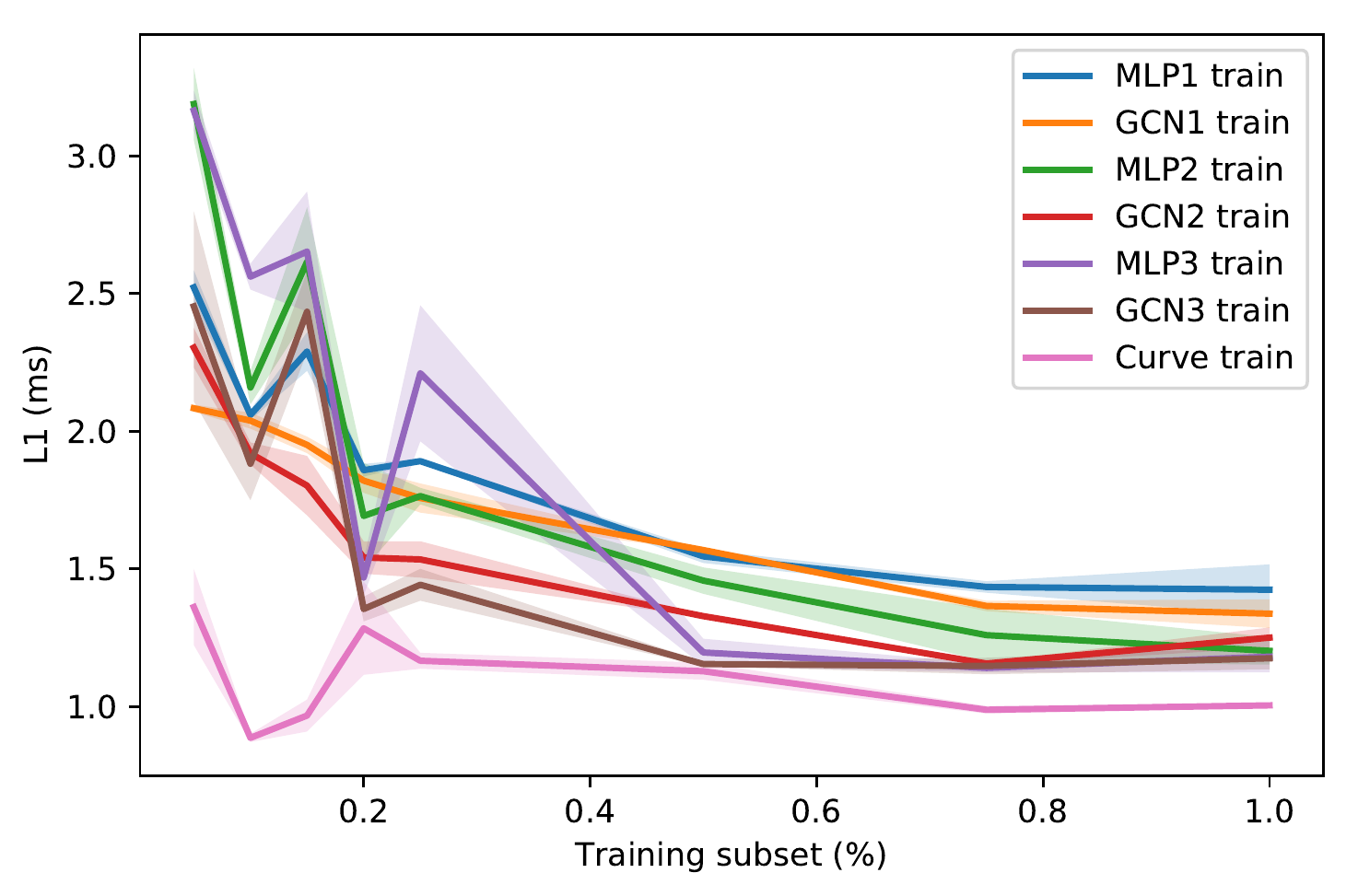}\quad
\includegraphics[width=0.45\textwidth]{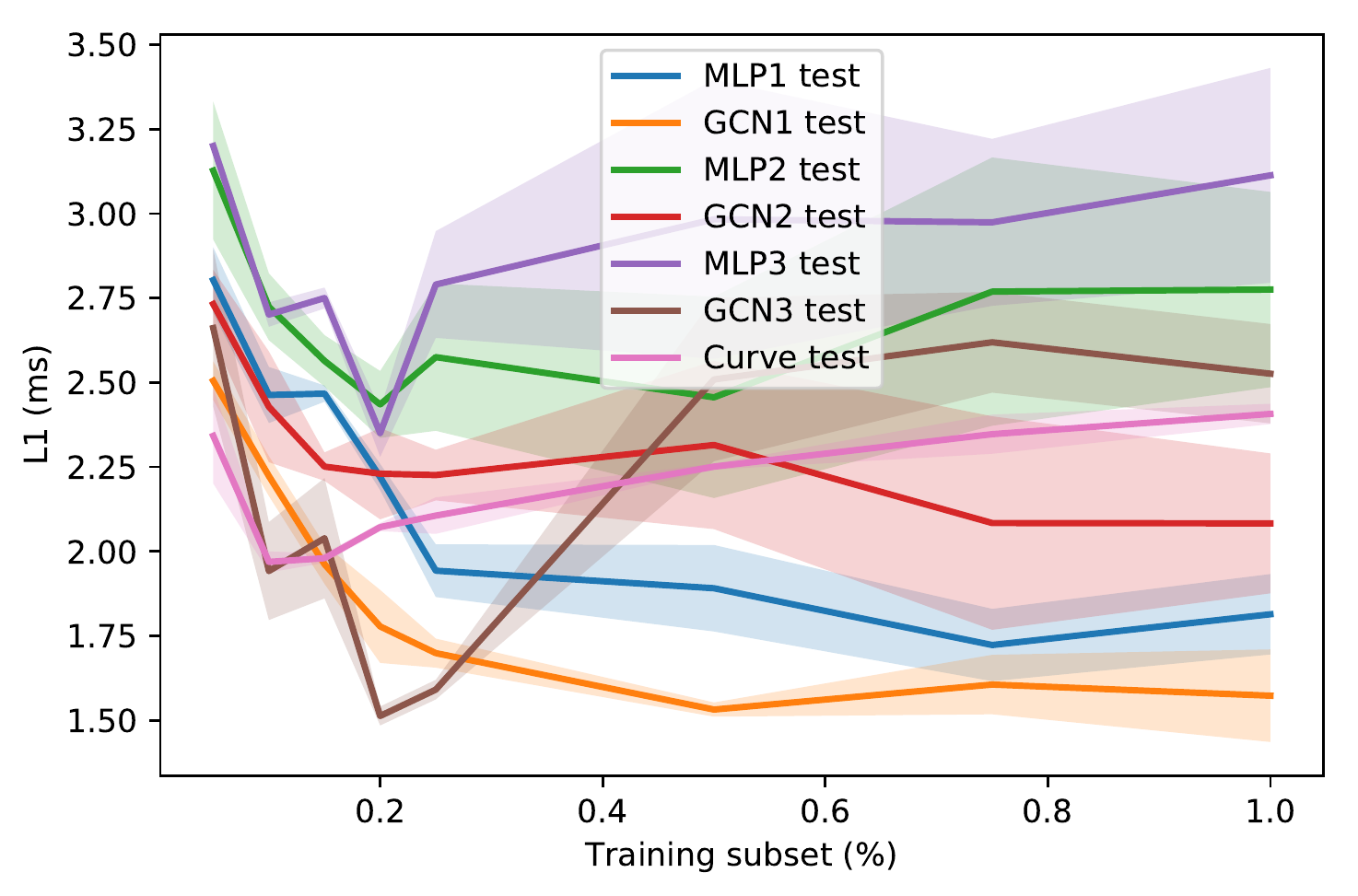}\quad
\caption{The final evaluation using $\ell_1$ loss: (\textit{left}) on training data, (\textit{right}) on test data.}
\label{fig:results_summary}
\end{figure}

\begin{figure}[!tbp]
\includegraphics[width=0.32\textwidth]{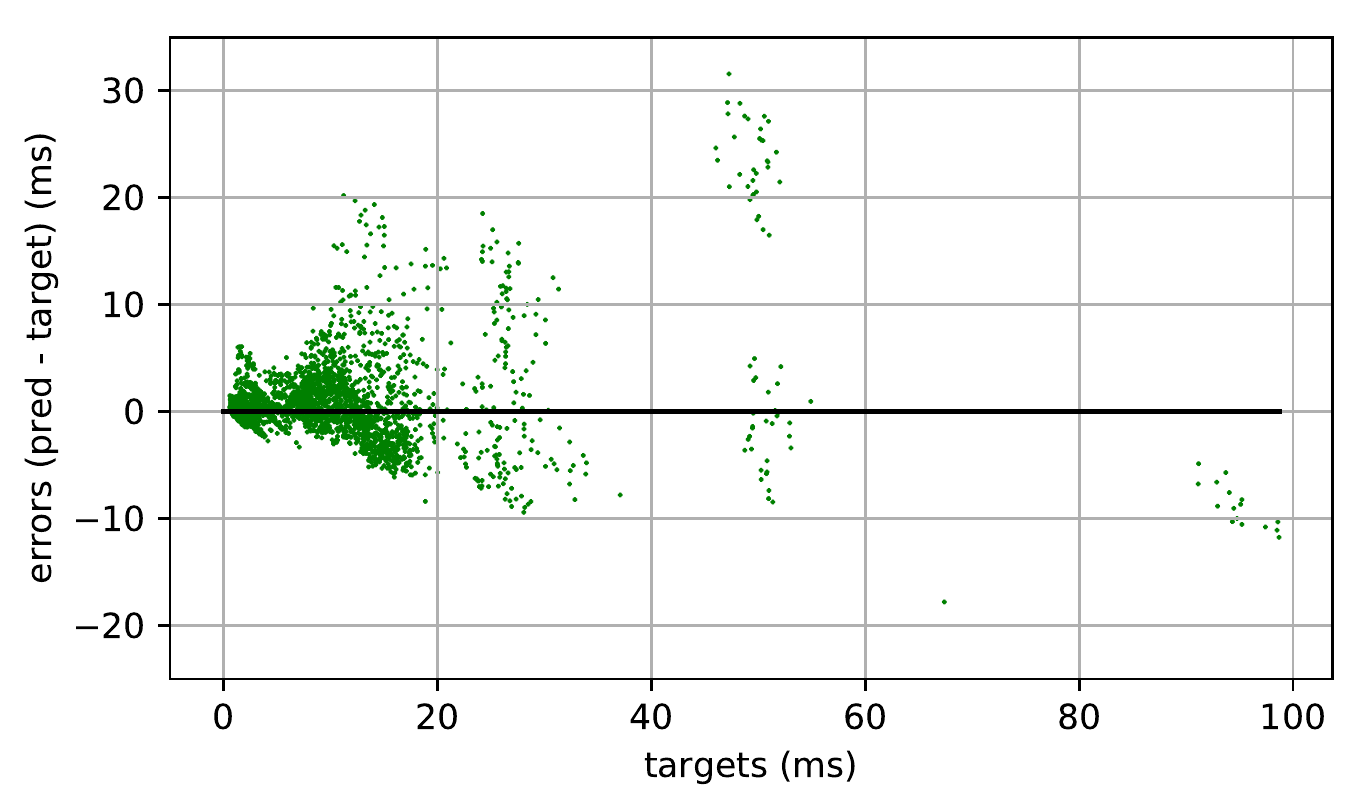}\quad
\includegraphics[width=0.32\textwidth]{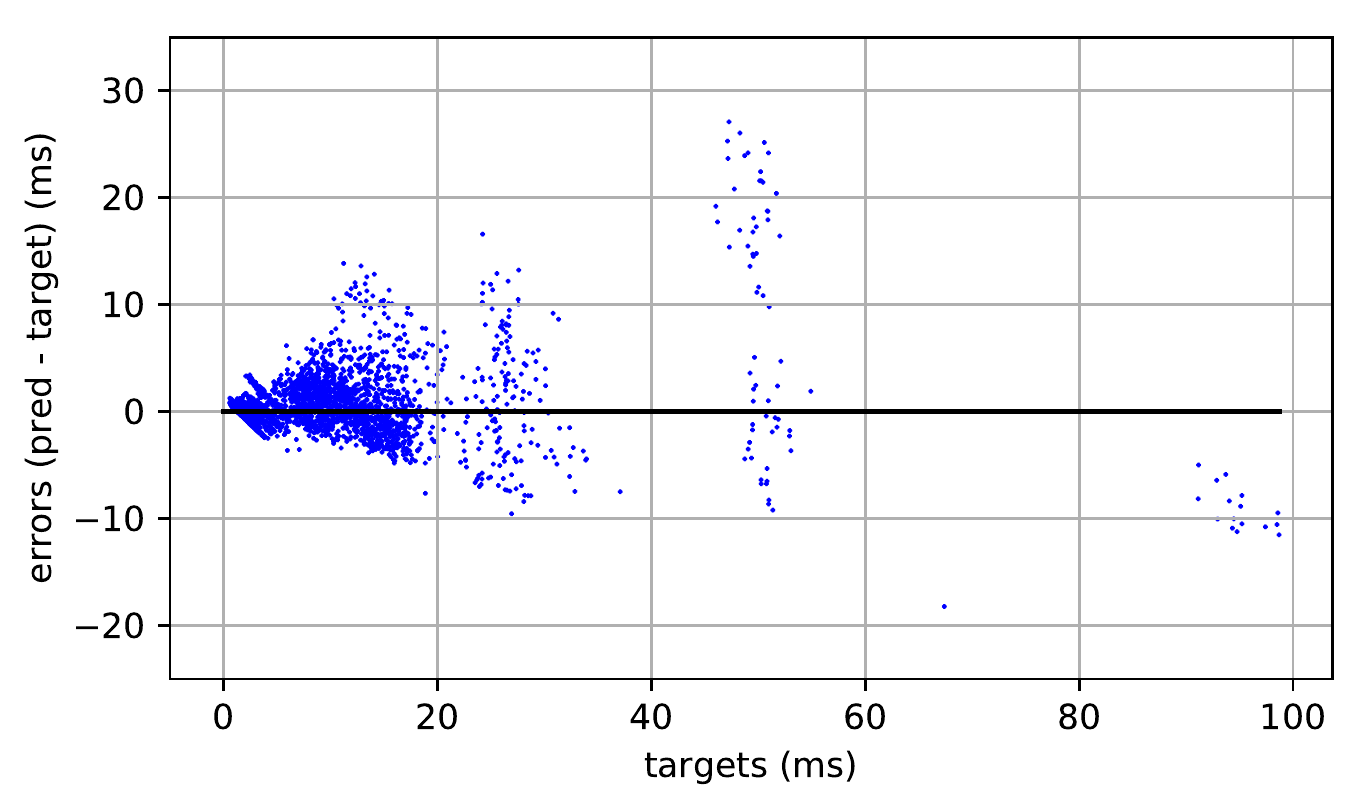}\quad
\includegraphics[width=0.32\textwidth]{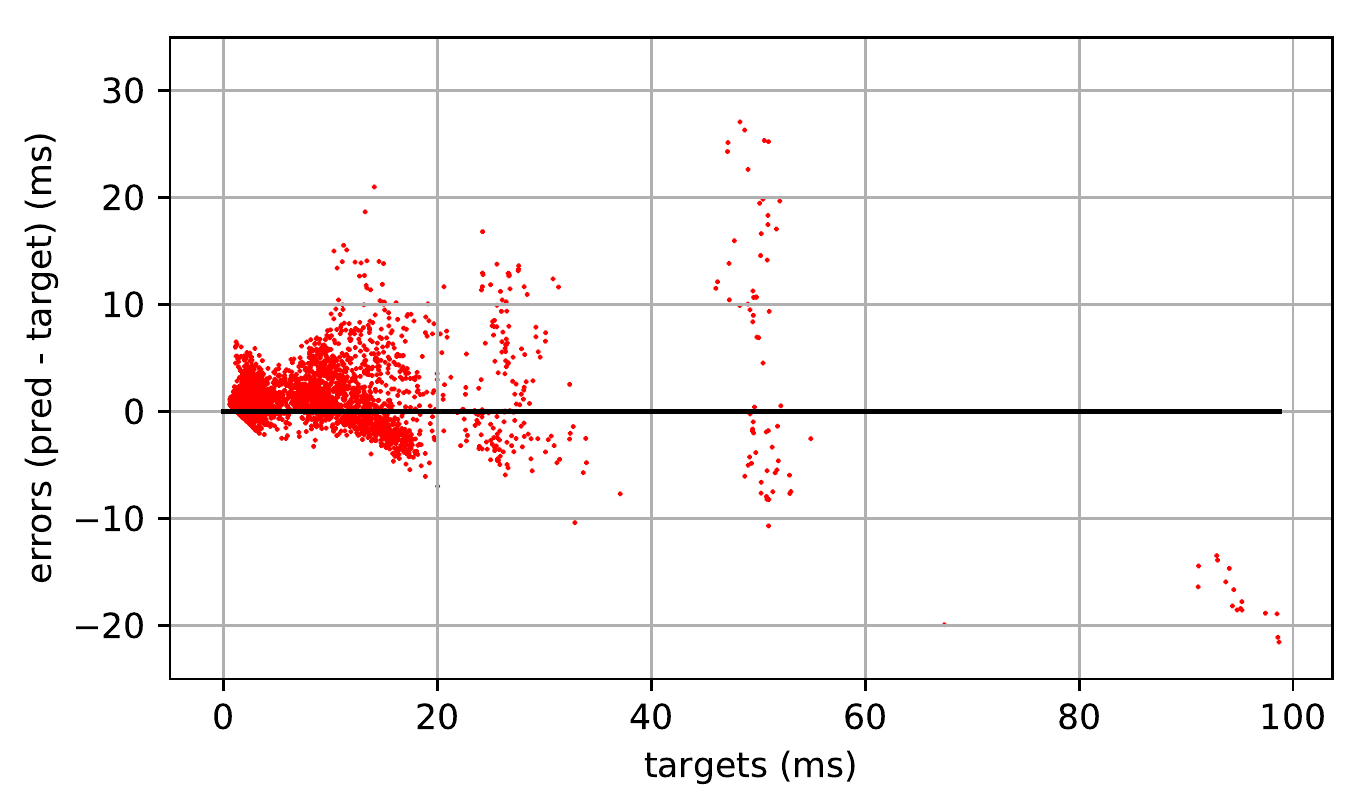}
\caption{Scatter plots of errors (y-axis) for targets (x-axis) on the test set where only $20\%$ of training data is available: (\textit{left}) MLP, (\textit{center}) GCN, (\textit{right}) Curve.}
\label{fig:scatter_error}
\end{figure}

%\begin{table}
%	\caption{Results for both experiments: $\ell_1$ is in milliseconds.}
%	\label{tab:cross-workload-generalization}
%\begin{center}
%	\begin{tabular}{c | c | c || p{10mm}| p{10mm} }
%		\multicolumn{3}{c}{Network architecture} & \multicolumn{2}{c}{Cross-workload} \\ 
%		\hline
%			$e(\cdot)$ & $g(\cdot)$ & $h(\cdot)$	& Val & Test \\
%		\hline
%			128-128 			& 128 		& 128-64 & 1.171 & 2.421 \\
%			128-128-128 		& identity	& 128-64 & 1.423 & 1.340 \\
%			128-128 			& 128-128 	& 128-64 & 1.423 & 1.340 \\
%			128-128-128-128 	& identity	& 128-64 & 1.423 & 1.340 \\
%		\hline
%			\multicolumn{3}{c|}{XGBoost}			 & x		 & x	
%	\end{tabular}
%\end{center}
%\end{table}

\subsection{Results and discussion} 

We present a final evaluation on $5$ cases with different training datasets in Figure \ref{fig:results_summary}.
We observe that GCN-based network architectures generalize to the unseen test data better than the models without the graph convolution component.
The Curve model performs on par with the best performing GCN for $5\%-15\%$ of training data, however, it generalizes poorly if more training data is available.
To further inspect the performance of the three methods we present differences between predictions and targets in Figure \ref{fig:scatter_error}.
A closer inspection reveals that the GCN makes smaller mistakes in general and, most importantly, for targets between $0$ and $20$.
Note that accurate prediction of target in $[0, 20]$ is critical since the goal of TVM is to identify a configuration with the smallest target execution time.
We conjecture that the main reason for this difference is that the GCN is able to better exploit local structure, as the MLP can only share information among nodes after the aggregation function has been applied.
Additionally, the results indicate that generalization to unseen workloads by training on similar workloads is possible to some extent, which shows that the found representations likely lend themselves well to transfer learning.
%It is noteworthy that despite the small dataset, train and test results are fairly close.

%--SECTION--
\section{Final remarks}
In this paper, we have presented a new dataset for learning the runtime of tensor program configurations represented as graphs. 
We provided three baselines, namely, a neural network with mean as the aggregation function, a GraphNN-based network that propagates information among nodes before the aggregation, and a neural network trained on context relational features representing a whole AST.
Additionally, we showed that the GraphNN-based approach allows to obtain competitive results to the fixed feature extractor.
We believe that this new dataset and the presented task will attract attention of the GraphNN community to the problem of learning tensor programs.

\subsubsection*{Acknowledgments}
We would like to thank Jinwon Lee for fruitful discussions and helpful remarks.

%\bibliography{compiler}
%\bibliographystyle{iclr2019_conference}

\newpage
%--SECTION--
\section*{Supplementary material}
\paragraph{Dataset: Extra details} To mitigate the variability of on-device measurements, each execution time is the average of 5 runs, where a run consists of taking 4 measurements and averaging these. Each edge in a graph is bidirectional (extraverted edges).

\paragraph{Implementation} We used PyTorch \citep{paszke2017automatic} to implement our framework and the Deep Graph Library package for GraphNNs \citep{dgl}.

\paragraph{Curve: \textit{context relation features}} Context relation features introduced in \citep{chen2018learning} are calculated with a hyperparameter determining the number of sample points along one dimension. In our experiments we used $20$ samples, the default value used for the TVM cost model.
 
\paragraph{Network architectures} Different models used in the paper with their corresponding labels are gathered in Table \ref{tab:architectures}.

\begin{table}[!hptb]
	\caption{A specification of architectures used in different approaches used in the paper.}
	\label{tab:architectures}
\begin{center}
	\begin{tabular}{c | c | c | c | c }
		 & \multicolumn{4}{c}{\textbf{Model architecture}} \\ 
		\hline
			\textbf{Label} &	$e(\cdot)$	& $g(\cdot)$	& $\rho(\cdot)$	& $h(\cdot)$	\\
		\hline
			MLP1	 & 128			& - 			& average		& 128-64 \\
			MLP2 & 128-128 		& - 			& average		& 128-64 \\
			MLP3 & 128-128-128	& - 			& average		& 128-64 \\
			\hline
			GCN1	 & -				& 128 		& average		& 128-64 \\
			GCN2 & 128 			& 128		& average		& 128-64 \\
			GCN3 & 128-128		& 128		& average		& 128-64 \\
			\hline
			Curve& -				& -			& -				& 128-64
	\end{tabular}
\end{center}
\end{table}

\paragraph{Training details} 
We use Adam optimizer with default hyperparameters, learning rate set to $1e$-$3$. 
For all experiments, we use patience-based learning rate decay, which multiplies the base learning rate by 0.9 if the validation loss did not decrease by 1\% (relative) in the last 6 epochs, with a minimum learning rate of $1e$-$6$. 
The maximum number of epochs is set $200$, and we stop training if the validation loss does not decrease in the last 20 epochs.

For the Curve model we checked $\ell_2$ regularization on weights (the weight decay) with different regularization coefficients ($0.01$ and $0.1$), however, we have noticed no significant improvement.

For GraphNNs, we use max-pooling for message passing, and only allow top-down messaging (from the root of the AST to its leaves), as this worked better in practice.
Similar to \citep{chen2018learning}, we use a learnable embedding to encode the node type as a vector of size 64.

\paragraph{Detailed results} 
We present detailed results of the experiment in Table \ref{tab:detailed_results}.

\begin{table}[!htp]
\centering
\caption{Detailed results of all experiments. An average and a standard error over $3$ runs are presented.}
\label{tab:detailed_results}
\vspace{2mm}
\resizebox{0.5\textwidth}{!}{
\begin{tabular}{c|c|c|c}
\textbf{Training subset} & \textbf{Model}     & \textbf{Train loss} & \textbf{Test loss} \\
\hline
\hline
0.05   & MLP1      & 2.52  $\pm$ 0.06 & 2.80 $\pm$ 0.10 \\
0.05   & MLP2      & 3.19  $\pm$ 0.13 & 3.13 $\pm$ 0.20 \\
0.05   & MLP3      & 3.16  $\pm$ 0.07 & 3.20 $\pm$ 0.01 \\
\hline
0.05   & GCN1      & 2.08  $\pm$ 0.01 & 2.51 $\pm$ 0.06 \\
0.05   & GCN2      & 2.30  $\pm$ 0.07 & 2.73 $\pm$ 0.10 \\
0.05   & GCN3      & 2.45  $\pm$ 0.35 & 2.66 $\pm$ 0.23 \\
\hline
0.05   & Curve     & 1.36  $\pm$ 0.14 & 2.34 $\pm$ 0.14 \\
\hline
\hline
0.1    & MLP1      & 2.06  $\pm$ 0.02 & 2.46 $\pm$ 0.08 \\
0.1    & MLP2      & 2.16  $\pm$ 0.06 & 2.72 $\pm$ 0.10 \\
0.1    & MLP3      & 2.56  $\pm$ 0.05 & 2.70 $\pm$ 0.04 \\
\hline
0.1    & GCN1      & 2.04  $\pm$ 0.03 & 2.22 $\pm$ 0.06 \\
0.1    & GCN2      & 1.92  $\pm$ 0.04 & 2.43 $\pm$ 0.16 \\
0.1    & GCN3      & 1.88  $\pm$ 0.13 & 1.94 $\pm$ 0.15 \\
\hline
0.1    & Curve     & 0.89  $\pm$ 0.02 & 1.97 $\pm$ 0.03 \\
\hline
\hline
0.15   & MLP1      & 2.29  $\pm$ 0.07 & 2.47 $\pm$ 0.02 \\
0.15   & MLP2      & 2.62  $\pm$ 0.20 & 2.56 $\pm$ 0.08 \\
0.15   & MLP3      & 2.65  $\pm$ 0.22 & 2.75 $\pm$ 0.03 \\
\hline
0.15   & GCN1      & 1.95  $\pm$ 0.03 & 1.96 $\pm$ 0.06 \\
0.15   & GCN2      & 1.80  $\pm$ 0.11 & 2.25 $\pm$ 0.04 \\
0.15   & GCN3      & 2.43  $\pm$ 0.15 & 2.04 $\pm$ 0.18 \\
\hline
0.15   & Curve     & 0.97  $\pm$ 0.06 & 1.98 $\pm$ 0.01 \\
\hline
\hline
0.2    & MLP1      & 1.86  $\pm$ 0.02 & 2.22 $\pm$ 0.04 \\
0.2    & MLP2      & 1.69  $\pm$ 0.19 & 2.44 $\pm$ 0.10 \\
0.2    & MLP3      & 1.47  $\pm$ 0.07 & 2.35 $\pm$ 0.07 \\
\hline
0.2    & GCN1      & 1.82  $\pm$ 0.04 & 1.78 $\pm$ 0.11 \\
0.2    & GCN2      & 1.54  $\pm$ 0.06 & 2.23 $\pm$ 0.14 \\
0.2    & GCN3      & 1.35  $\pm$ 0.05 & 1.51 $\pm$ 0.03 \\
\hline
0.2    & Curve     & 1.28  $\pm$ 0.17 & 2.07 $\pm$ 0.01 \\
\hline
\hline
0.25   & MLP1      & 1.89  $\pm$ 0.01 & 1.94 $\pm$ 0.08 \\
0.25   & MLP2      & 1.76  $\pm$ 0.03 & 2.58 $\pm$ 0.22 \\
0.25   & MLP3      & 2.21  $\pm$ 0.25 & 2.79 $\pm$ 0.16 \\
\hline
0.25   & GCN1      & 1.76  $\pm$ 0.05 & 1.70 $\pm$ 0.04 \\
0.25   & GCN2      & 1.53  $\pm$ 0.07 & 2.23 $\pm$ 0.08 \\
0.25   & GCN3      & 1.44  $\pm$ 0.06 & 1.59 $\pm$ 0.03 \\
\hline
0.25   & Curve     & 1.17  $\pm$ 0.03 & 2.11 $\pm$ 0.05 \\
\hline
\hline
0.5    & MLP1      & 1.55  $\pm$ 0.02 & 1.89 $\pm$ 0.13 \\
0.5    & MLP2      & 1.46  $\pm$ 0.05 & 2.46 $\pm$ 0.30 \\
0.5    & MLP3      & 1.20  $\pm$ 0.05 & 2.98 $\pm$ 0.41 \\
\hline
0.5    & GCN1      & 1.57  $\pm$ 0.01 & 1.53 $\pm$ 0.02 \\
0.5    & GCN2      & 1.33  $\pm$ 0.00 & 2.32 $\pm$ 0.25 \\
0.5    & GCN3      & 1.15  $\pm$ 0.01 & 2.51 $\pm$ 0.24 \\
\hline
0.5    & Curve     & 1.13  $\pm$ 0.03 & 2.25 $\pm$ 0.01 \\
\hline
\hline
0.75   & MLP1      & 1.43  $\pm$ 0.02 & 1.72 $\pm$ 0.11 \\
0.75   & MLP2      & 1.26  $\pm$ 0.10 & 2.77 $\pm$ 0.40 \\
0.75   & MLP3      & 1.14  $\pm$ 0.02 & 2.97 $\pm$ 0.25 \\
\hline
0.75   & GCN1      & 1.37  $\pm$ 0.02 & 1.61 $\pm$ 0.09 \\
0.75   & GCN2      & 1.16  $\pm$ 0.01 & 2.08 $\pm$ 0.32 \\
0.75   & GCN3      & 1.15  $\pm$ 0.03 & 2.62 $\pm$ 0.15 \\
\hline
0.75   & Curve     & 0.99  $\pm$ 0.01 & 2.35 $\pm$ 0.06 \\
\hline
\hline
1      & MLP1      & 1.42  $\pm$ 0.09 & 1.81 $\pm$ 0.12 \\
1      & MLP2      & 1.20  $\pm$ 0.05 & 2.78 $\pm$ 0.29 \\
1      & MLP3      & 1.18  $\pm$ 0.06 & 3.11 $\pm$ 0.32 \\
\hline
1      & GCN1      & 1.34  $\pm$ 0.05 & 1.57 $\pm$ 0.14 \\
1      & GCN2      & 1.25  $\pm$ 0.04 & 2.08 $\pm$ 0.21 \\
1      & GCN3      & 1.18  $\pm$ 0.04 & 2.53 $\pm$ 0.15 \\
\hline
1      & Curve     & 1.00  $\pm$ 0.01 & 2.41 $\pm$ 0.03
\end{tabular}
}
\end{table}
\end{document}